\pdfoutput=1

\documentclass[11pt]{article}

\usepackage{ACL2023}
\usepackage{times}
\usepackage{latexsym}

\usepackage[T1]{fontenc}

\usepackage[utf8]{inputenc}

\usepackage{microtype}

\usepackage{inconsolata}

\usepackage{amsfonts}
\usepackage{amsmath}
\usepackage{enumitem}
\usepackage{graphicx}
\usepackage{kantlipsum}
\usepackage{wrapfig}
\usepackage{comment}

\newcommand{\be}{\begin{equation}}
	\newcommand{\ee}{\end{equation}}
\newcommand{\bes}{\begin{equation*}}
	\newcommand{\ees}{\end{equation*}}

\usepackage{array}
\newcommand{\head}[2]{\multicolumn{1}{>{\centering\arraybackslash}p{#1}}{\textbf{#2}}}

\usepackage{ragged2e}

\usepackage[utf8]{inputenc} 
\usepackage[T1]{fontenc}    
\usepackage{hyperref}       
\usepackage{url}            
\usepackage{booktabs}       
\usepackage{amsfonts}       
\usepackage{nicefrac}       
\usepackage{microtype}      
\usepackage{times}
\usepackage{latexsym}
\usepackage[flushleft]{threeparttable}

\title{Measuring and Mitigating Local Instability in Deep Neural Networks}

\author{
        Arghya Datta$^{\dag}$,  
        Subhrangshu Nandi$^{\dag}$, 
        Jingcheng Xu$^{\dag}$\thanks{$^{*}$ Work done as an intern at Amazon Alexa}, 
         Greg Ver Steeg, \\ 
        {\bf He Xie},  
        {\bf Anoop Kumar}, 
        {\bf Aram Galstyan}, \\ 
        Amazon Alexa \\ 
        Seattle, WA, USA \\ 
        \texttt{\{argdatta, subhrn, gssteeg, hexie, anooamzn, argalsty\} @amazon.com} \\ 
        \texttt{\{xjc\}@stat.wisc.edu} \\
        $^{\dag}$ Equal Contribution, alphabetical order \\
        To appear in \textit{Findings of Association for Computational Linguistics(ACL), 2023}
        }

\begin{document}
\maketitle

\vspace{-0.5cm}
\begin{abstract}
\vspace{-0.25cm}

Deep Neural Networks (DNNs) are becoming integral components of real world services relied upon by millions of users. Unfortunately, architects of these systems can find it difficult to ensure reliable performance as irrelevant details like random initialization can unexpectedly change the outputs of a trained system with potentially disastrous consequences. 
We formulate the model stability problem by studying how the predictions of a model change, even when it is retrained on the same data, as a consequence of stochasticity in the training process. 
For Natural Language Understanding (NLU) tasks, we find instability in predictions for a significant fraction of queries. We formulate principled metrics, like per-sample ``label entropy'' across training runs or within a single training run, to quantify this phenomenon. 
Intriguingly, we find that unstable predictions do not appear at random, but rather appear to be clustered in data-specific ways. 
We study data-agnostic regularization methods to improve stability and propose new data-centric methods that exploit our local stability estimates. 
We find that our localized data-specific mitigation strategy dramatically outperforms data-agnostic methods, and comes within 90\% of the gold standard, achieved by ensembling, at a fraction of the computational cost. 


\end{abstract}

\vspace{-0.25cm}
\section{Introduction}
\vspace{-0.15cm}
When training large deep neural networks on the same data and hyperparameters can lead to many distinct solutions with similar loss, we say the model is \emph{underspecified}~\cite{Damour2022}.  One tangible manifestation of underspecification  is that a model prediction on a single data point can change across different training runs, without any change in the training data or hyperparameter settings, due to stochasticity in the training procedure. This extreme sensitivity of model output, which has been termed as \textit{model variance/instability} or \textit{model jitter/churn}~\cite{hidey2022reducing,milani2016launch}, is highly undesirable as it prohibits comparing models across different experiments~\cite{dodge-etal-2019-show}. We refer to this problem as \emph{local instability} \footnote{We use \emph{local instability} to mean \emph{local model instability}}, a term that highlights our focus on the non-uniformity of instability across data points. 
Local instability can lead to highly undesirable consequences for deployed industrial systems, as it can cause inconsistent model behavior across time, eroding trust on AI systems \cite{dodge2020fine,d2020underspecification}. The problem is further exacerbated by the fact that industry models are typically more complex and trained on diverse datasets with potentially higher proportion of noise. 

\begin{table}[!ht]
\small
\centering
\begin{tabular}{p{2.2cm}|p{1.8cm}|p{2.3cm}}
\midrule
\head{2.2cm}{Utterance (gold label)} & \head{1.8cm}{$\hat{p}_{[min-max]}$, ${\sigma_m(\hat{p})}$} & \head{2.3cm}{Label predictions over 50 runs} \\
\hline
\RaggedRight{\textcolor{green}{funny joke (general)}} & [0.98-0.99], 0.003 (low) & general:50 \\
\hline
\RaggedRight{\textcolor{red}{start house cleanup (IOT)}} & [0.002-0.97], 0.17 (high)& \RaggedRight{lists:26, IOT:6, general:6, play:5, news:3, social:1, calendar:1}\\
\hline
\RaggedRight{\textcolor{red}{search for gluten free menus (cooking)}} & [0.002-0.693], 0.06 (low) & \RaggedRight{lists:28, takeaway:18, social:1, music:1, cooking:1, play:1} \\
\bottomrule
\end{tabular}
\caption{Utterances from Massive data show different predictions over 50 model runs with different seeds. $\hat{p}$ is the prediction score on gold labels and $\sigma_m$ is the standard deviation over \textit{m}ultiple model outputs $\hat{p_1}, \dots, \hat{p_{50}}$. For example, \textit{start house cleanup} with gold label \textit{IOT} is predicted to label \textit{lists} 26 out of the 50 model runs. Its prediction score on \textit{IOT} ranges between 0.002 and 0.97. \textcolor{green}{green}: low variability, predictions match gold label, \textcolor{red}{red}: high predicted label switching}
\label{table:intro_example}
\end{table}

Table  \ref{table:intro_example} shows examples of local instability for a domain classification problem, where we used a pre-trained language model DistilBERT \cite{sanh2019distilbert} to train 50 independent classifiers (with random initial conditions) on Massive dataset ~\cite{fitzgerald2022massive}. It shows that a validation set utterance \textit{start house cleanup} with gold label \textit{IOT}  gets assigned seven different predicted labels over the 50 runs, with the predicted confidence on gold label $\hat{p}$ ranging between 0.002 and 0.97, with high $\sigma_m$ (the standard deviation of $\{\hat{p_i}\}_{i=1}^{50}$) of 0.17. In comparison, \textit{search for gluten free menus} gets 6 different predicted labels over 50 runs, with a relatively low $\sigma_m$ of 0.06. 
The differences in stability across examples demonstrates that the phenomenon is localized to certain data points. See Figures \ref{fig:Massive_mu_vs_sigma} and \ref{fig:Clinc150mu_vs_sigma} in Appendix. Examples in table \ref{table:intro_example} also highlight that variability in confidence is not perfectly aligned with stability of predictions. 

\paragraph{Measuring Local Model Instability} While detecting and quantifying local instability across multiple runs is trivial for toy problems, it becomes infeasible with much larger industrial datasets. ~\cite{swayamdipta2020dataset} suggested to use single-run training dynamics to estimate the variance  in prediction scores over multiple epochs. However, as shown in Table~ \ref{table:intro_example} low prediction variance does not always lead to less label switching, which is the defining feature of local instability. Instead, here we introduce {\em label switching entropy} as a new metric for characterizing local instability. Furthermore, we demonstrate that label switching entropy calculated over training epochs of a single run is a good proxy for label switching over multiple runs, so that data points with high prediction instability over time also exhibit high instability across training runs. 

\paragraph{Mitigating Local Model Instability} One straightforward strategy of mitigating local instability is to train an ensemble of $n$ models and average their weights or their predictions. Unfortunately, ensembling neural networks such as large language models is often computationally infeasible in practice, as it requires multiplying both the training cost and the test time inference cost by a factor of $n$. Therefore, we propose and compare more economical options for mitigating local instability. 


Here we propose a more efficient smoothing-based approach where we train just two models. The first (teacher) model is trained using the one-hot encoded gold labels as the target. Once the model has converged and is no longer in the transient learning regime (after $N$ training or optimization steps), we compute the temporal average predicted probability vector over $K$ classes after each optimization step, which is then adjusted by temperature $T$ to obtain the smoothed predicted probability vector. A student model is then trained using these ``soft'' labels instead of the one-hot encoded gold labels. We call this Temporal Guided Temperature Scaled Smoothing (TGTSS). TGTSS allows local mitigation of local instability as each datapoint is trained to its unique label in the student model. In contrast to existing methods such stochastic weight averaging \cite{Izmailov2018AveragingWL} or regularizing options such as adding L2-penalty, TGTSS significantly outperforms existing methods and reaches within 90\% of the gold standard of ensemble averaging. 


We summarize our  contributions as follows:
\begin{itemize}
\item We propose a new measure of local instability that is computationally efficient and descriptive of actual prediction changes.
\item We introduce a data-centric strategy to mitigate local instability by leveraging temporally guided label smoothing.
\item We conduct extensive experiments with two public datasets and demonstrate the effectiveness of the proposed mitigation strategy compared to existing baselines. 
\end{itemize}

\vspace{-0.25cm}
\section{Related work}
\vspace{-0.15cm}
\label{sec:related_work}

Sophisticated, real-world applications of Deep Neural Networks (DNNs) introduce challenges that require going beyond a myopic focus on accuracy. 
Uncertainty estimation is increasingly important for deciding when a DNN's prediction should be trusted, by designing calibrated confidence measures that may even account for differences between training and test data~\cite{nado2021uncertainty}. 
Progress on uncertainty estimation is largely orthogonal to another critical goal for many engineered systems: \emph{consistency} and \emph{reliability}. 
Will a system that works for a particular task today continue to work in the same way tomorrow? 
One reason for inconsistent performance in real-world systems is that even if a system is re-trained with the same data, predictions may significantly change, a phenomenon that has been called model \emph{churn}\cite{milani2016launch}.
The reason for this variability is that neural networks are under-specified~\cite{underspecification}, in the sense that there are many different neural networks that have nearly equivalent average performance for the target task. 
While randomness could be trivially removed by fixing seeds, in practice tiny changes to data will still significantly alter stochasticity and results. We will explore the case of altering training data in future studies. Studying how stochasticity affects model churn addresses a key obstacle in re-training engineered systems while maintaining consistency with previous results. 

The most common thread for reducing model churn focuses on adding constraints to a system so that predictions for re-trained system match some reference model. This can be accomplished by adding hard constraints~\cite{cotter2019optimization} or distillation~\cite{milani2016launch,jiang2021churn, bhojanapalli2021reproducibility}. 

We adopt a subtly different goal which is to train at the outset in a way that reduces variability in predictions due to stochasticity in training.  \cite{hidey2022reducing} suggest a co-distillation procedure to achieve this. Label smoothing, which reduces over-confidence~\cite{muller2019does}, has also been suggested to reduce variance, with a local smoothing approach to reduce model churn appearing in \cite{bahri2021locally}. 

A distinctive feature of our approach is a focus on how properties of the data lead to instability. Inspired by dataset cartography~\cite{swayamdipta2020dataset} which explored variance in predictions over time during training of a single model, we investigate how different data points vary in predictions across training runs. Non-trivial patterns emerge, and we use sample-specific instability to motivate a new approach to reducing model churn. 

Our work draws connections between model stability and recent tractable approximations for Bayesian learning~\cite{Izmailov2018AveragingWL,swag}. Recent Bayesian learning work focuses on the benefits of Bayesian model ensembling for confidence calibration, but an optimal Bayesian ensemble would also be stable. Bayesian approximations exploit the fact that SGD training dynamics approximate MCMC sampling, and therefore samples of models over a single training run can approximate samples of models across training runs, although not perfectly~\cite{fort2019deep,wenzel2020good,izmailov2021bayesian}. We study connections between prediction variability within a training run and across training runs, and use this connection to devise practical metrics and mitigation strategies. 

Similar to BANNs ~\cite{furlanello_bann}, our teacher and corresponding student models use the same model architecture with same no. of parameters rather than using a high-capacity teacher model, however, unlike BANNS, our work is geared towards addressing model instability. Architecturally, our methodology (TGTSS) uses a temperature scaled temporally smoothed vector that is obtained from the last \textit{N} checkpoints from the teacher model instead of the finalized teacher model and not use the annotated labels for the utterances.



\vspace{-0.25cm}
\section{Model instability measurement}
\vspace{-0.15cm}
\label{sec:metrics}
The examples in Table~\ref{table:intro_example} show that re-training a model with different random seeds can lead to wildly different predictions. The variance of predictions across models, $\sigma_m^2$, is intuitive, but is expensive to compute and does not necessarily align with user experience since changes in confidence may not change predictions. A changed prediction, on the other hand, may break functionality that users had come to rely on. Hence we want to include a metric which measures how often predictions change. 

\begin{figure}[htb!]
 \centering
  \includegraphics[width=\columnwidth]{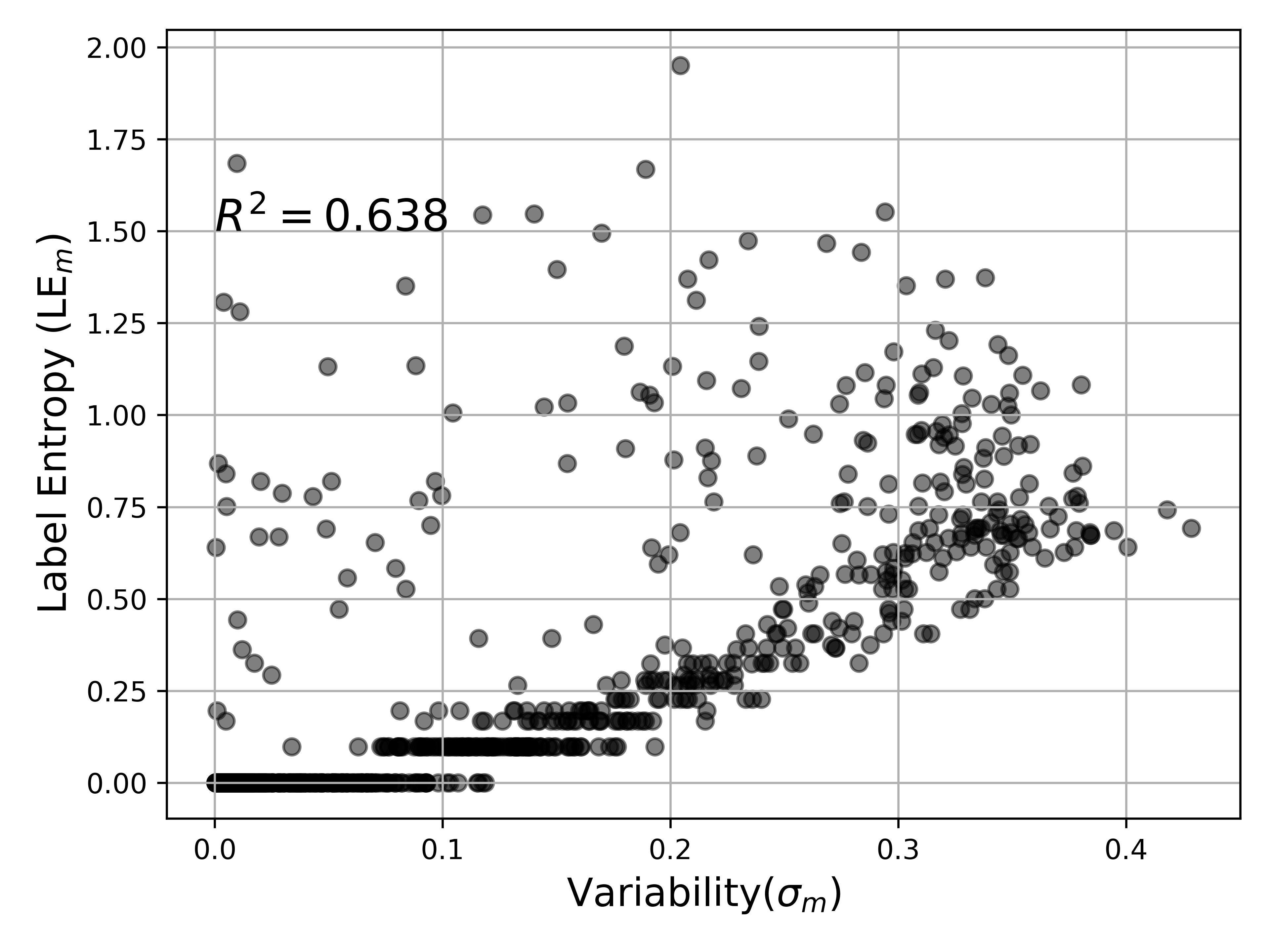}
    \caption{$LE_m$ vs $\sigma_m$ for Massive dataset shows a strong linear relationship. Each data point is an utterance with $LE_m^{(i)}$ vs $\sigma_m^{(i)}$ values.}
    \label{fig:lem_vs_std}
\vspace{-0.25cm}
\end{figure}

Therefore, we propose to study the label switching entropy. 
Given a setup with training data $\{x_i,y_i\} \ \in X$ where $X$ are utterances, $y \in \{1,...,K\}$ are the corresponding gold labels, the \underline{m}ulti-run \underline{L}abel \underline{E}ntropy ($LE_m$) over $N$ independent runs for an utterance $x_i$ can be computed as, 
\begin{equation}
 LE_m^{(i)} =  {\sum_{k=1}^{K}} - \frac{n_k^{(i)}}{N}\log(\frac{n_k^{(i)}}{N}) \label{eq:1}
\end{equation}
where, $n_k$ is the number of times utterance $i$ was predicted to be in class $k$ across $N$ models trained with different random seeds. For example, if an utterance gets labeled to three classes A, B and C for 90\%, 5\% and 5\% of the time respectively, then its multi-run label entropy ($LE_m^{(i)}$) will be  $ -(0.9*\log (0.9) + 0.05 * \log0.05 + 0.05 \log 0.05) = 0.39$. 
Similarly, an utterance that is consistently predicted to belong to one class over $N$ runs will have a $LE_m^{(i)}$ of 0 (even if it is consistently put in the \emph{wrong} class). We can compute the overall $LE_m$ by averaging $LE_m^{(i)}$ for all the utterances. Empirically, we also observe a relatively strong linear relationship between $LE_m$ and $\sigma_m$ (Figure ~\ref{fig:lem_vs_std}).

Since computing $LE_m$ is computationally expensive due to training $N$ independent models, we propose using \textit{\underline{s}ingle-run \underline{L}abel \underline{E}ntropy} ($LE_s$) that can be computed over a single model run. Mathematically, the formula for label entropy stays consistent for both multi-run and single-run, however, $LE_s$ is computed across different model checkpoints. In our analyses, we computed $LE_s$ by accumulating the predicted class after each optimization step whereas $LE_m$ was computed by accumulating the final predicted class across $N$ models on the validation set.

\begin{figure}[htb!]
 \centering
  \includegraphics[width=\columnwidth]{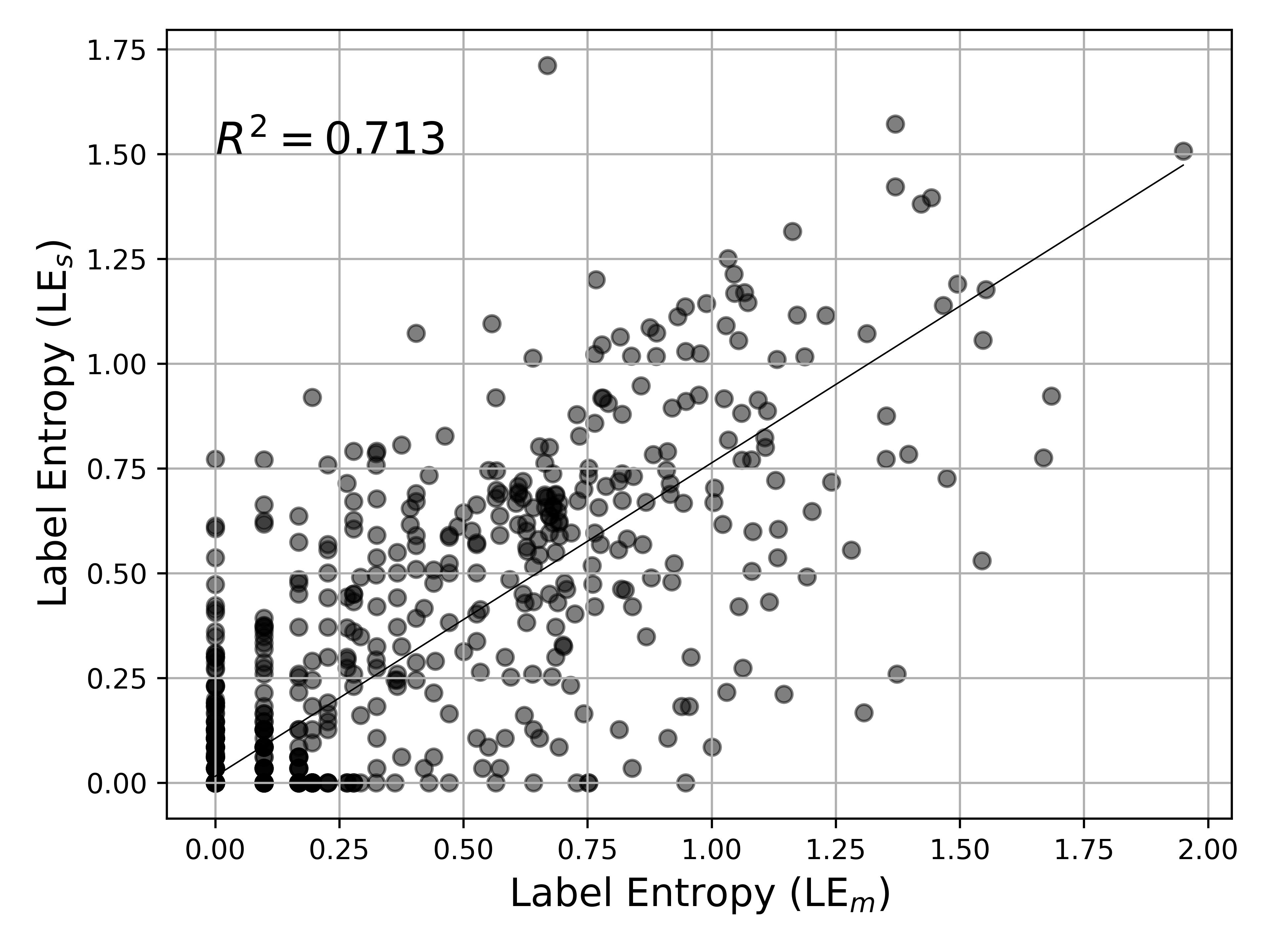}
    \caption{$LE_s$ vs $LE_m$ for Massive dataset shows a strong linear relationship. Each data point is an utterance with $LE_s^{(i)}$ vs $LE_m^{(i)}$ values. Zero entropy corresponds to utterances with confidence scores close to 1 for a class with very low variability.}
    \label{fig:les_vs_lem}
\vspace{-0.25cm}
\end{figure}

Empirically, we found that there exists a strong linear relationship between $LE_s$ and $LE_m$ (Figure ~\ref{fig:les_vs_lem}). This demonstrates that utterances that suffer from local instability across multiple independent runs exhibit similar instability across multiple optimization steps for a single model. This finding supports our hypothesis that $LE_s$ is a suitable proxy for $LE_m$ in real world production settings for NLU systems.


\vspace{-0.25cm}
\section{Model instability mitigation}
\vspace{-0.15cm}
\label{sec:mitigation}

In our study, we have explored 3 baseline mitigation strategies to address model instability: ensembling, stochastic weight averaging (SWA) and uniform label smoothing. These methodologies have been used in numerous other works to improve generalization as well as predictive accuracy across a diverse range of applications. Performance of the ensembling strategy serves as our upper bound in reducing model instability. We propose a novel model instability mitigation strategy, temporal guided temperature scaled label smoothing, that is able to recover 90\% of the reduction in model instability as ensembling at a fraction of model training time and computational cost. We describe all the mitigation strategies below.

\subsection{Ensemble averaging and regularizing}
In this setting, we trained \textit{N} independent models, initialized with different random seeds, using the standard cross-entropy loss, computed between the ground truth labels and the predicted probability vector. For every utterance in the test set, we recorded the mean predicted probability of the gold label, the predicted label and our proposed local instability metric, label entropy, across \textit{N} models. We also trained another baseline by leveraging $L2$ regularization. No other mitigation strategies were used in the process since our aim was to emulate the current model training scenario in natural language understanding(NLU) production settings. 

\subsection{Stochastic Weight Averaging}
Stochastic weight averaging(SWA) ~\cite{Izmailov2018AveragingWL} is a simple yet effective model training methodology that improves generalization performance in deep learning networks. SWA performs an uniform average of the weights traversed by the stochastic gradient descent based optimization algorithms with a modified learning rate. In our implementation, we equally averaged the weights at the end of the last two training epochs. We also explored equal averaging of weights from two randomly selected epochs out of the final 3 epochs but that strategy did not yield better results. We left the work of using a modified learning rate to a future study with a significantly larger training dataset.

\subsection{Label smoothing}
Label smoothing ~\cite{szegedy2016rethinking} is a popular technique to improve performance, robustness and calibration in deep learning models. Instead of using ``hard'' one-hot labels when computing the cross-entropy loss with the model predictions, label smoothing introduces ``soft'' labels that are essentially a weighted mixture of one-hot labels with the uniform distribution. For utterances $\{x_i,y_i\}$ where $y \in \{1,...,K\}$ for $K$ classes, the new "soft" label is given by $y^{LS} = (1-\alpha)*y +\alpha/K$ where $\alpha$ is the label smoothing parameter. The "soft" labels are then used in the softmax cross-entropy loss. 

\subsection{Ensemble baseline}
To obtain consistent predictions with low local instability, ensembling is often utilized as the default mitigation strategy. Given a problem setup with training data $\{x_i,y_i\} \ \in X$ where $X$ are utterances, $y \in \{1,...,K\}$ are the corresponding gold labels, then intuitively, ensembling over \textit{N} independent models,where \textit{N} is sufficiently large, will converge to the average predicted probability by the law of large numbers. Hence, using a sufficiently large ensemble of independently trained models would give stable predictions in general.

In our study, we used ensembling to aggregate (uniform average) predictions for each utterance across $N$ independently trained models. Each model was trained using the softmax cross-entropy loss between the predicted logit $z_i$ over $K$ classes and the one-hot encoded vector representing the gold label. For an utterance $x_i$, the uniform average predicted probability vector $\Bar{p}_i$ across $N$ models over all class $K$ (softmax probability vector of size $k = (1,K)$) is adjusted by a temperature $T$, to obtain the smoothed predicted probability vector $q_i$:

\vspace{-0.25cm}

\begin{equation}
 q_i = \frac{\Bar{p_i}^{T}}{\sum_{k=1}^{K} \Bar{p_k}^{T}} \label{eq:2}
\end{equation}


The temperature $T$ can be used to control the entropy of the distribution. The smoothed probability vector $q$ is now used as the "soft" labels to train a model instead of the "hard" one hot encoded gold labels and the resultant model is robust to local instability. One challenge for ensembling is that it requires training, storing and running inference on a large number of models which is often infeasible for large scale NLU systems. 

\subsection{Temporal guided temperature scaled smoothing (TGTSS)}
\label{sec:tgtss}
Since ensembling is infeasible for large models in practice, we propose temporal guided label smoothing that does not require training large ensembles to compute the soft labels. 

In this setup, we train a pair of models as opposed to training a large ensemble of models. The first (teacher) model is trained using the one-hot encoded gold labels as the target. Once the model has converged and is no longer in the transient training state (after $N$ training or optimization steps), we compute the uniform average predicted probability vector ($\Bar{p_i}$) after each optimization step of the model, which is then adjusted by temperature $T$ to obtain the smoothed predicted probability vector $q_i$ using \eqref{eq:2}. A suitable $N$ can be chosen by looking at the cross-entropy loss curve for the validation dataset. The second (student) model is now trained using $q_i$ as the "soft" label instead of the one-hot encoded gold labels.

The significant advantage of TGTSS over ensembling is that it does not require 
training, storing, or inferring over large ensembles. A key feature of TGTSS is that it uniformly averages predictions over numerous training steps instead of averaging predictions over numerous independent models. This saves the cost of training multiple models. Moreover, we never need to store multiple models for TGTSS since we can store a running average of the predictions over time. 
Finally, at inference time we only need to call a single model (the trained student model), as opposed to $N$ models for the ensemble.

\vspace{-0.25cm}
\section{Experimental setup and results for mitigation}
    \vspace{-0.15cm}
\label{sec:experiments}

\subsection{Base model architecture}

For all our experiments, we used DistilBERT~\cite{sanh2019distilbert} as the pre-trained language model. We used the implementation of \textit{DistilBERT-base-uncased} from the \textit{Huggingface} library by leveraging \textit{AutoModelForSequenceClassification}. The pre-trained language model is then fine-tuned on the benchmark datasets by using the training set. DistilBERT is a widely used pre-trained language model that is currently used in production in many large scale NLU systems. One key advantage of using DistilBERT is that it is able to recover more than 90\% performance of the larger \textit{BERT-base-uncased} model while using 40\% less parameters on the GLUE language understanding benchmark~\cite{wang2018glue}. Using other BERT models as the pre-trained language model was outside the scope of this study.

\subsection{Datasets}
To study local instability and compare different mitigation strategies, we used two open source benchmark datasets (Table~\ref{table:dataset_stats}): Massive and Clinc150. 
\begin{itemize}
    \item Massive: Massive~\cite{fitzgerald2022massive} dataset is an open source multilingual NLU dataset from Amazon Alexa NLU system consisting of 1 million labeled utterances spanning 51 languages. For our experiments, we only used the \textit{en-US} domain utterances for domain classification task across 18 domains (alarm, audio, general, music, recommendation, etc.). 
    \item Clinc150 DialoGLUE: Clinc150~\cite{larson-etal-2019-evaluation} is an open source dataset from DialoGLUE~\cite{Mehri2020DialoGLUEAN}, a conversational AI benchmark collection. We utilized Clinc150 for intent classification task across 150 intents (translate, transfer, timezone, taxes, etc).  
\end{itemize}
 
\begin{table}[!ht]
\centering
\small
\begin{tabular}{p{2.4cm}|p{1.6cm}|p{2.2cm}}
\toprule
\textbf{Attribute}      & \textbf{MASSIVE}  & \textbf{CLINC150}  \\\midrule
Source                  & Amazon Alexa AI & DialoGLUE                \\\midrule
Domains & \RaggedLeft{18} & \RaggedLeft{-}  \\
Intents & \RaggedLeft{60} & \RaggedLeft{150} \\
\midrule
Train & \RaggedLeft{11,514} & \RaggedLeft{15,000}  \\
Holdout(Unseen) & \RaggedLeft{2974} & \RaggedLeft{3,000} \\\midrule
Balanced? & \RaggedLeft{No.} & \RaggedLeft{Yes. 100 per intent} \\\midrule
Classification task     & \RaggedLeft{Domain} & \RaggedLeft{Intent}                        \\\bottomrule
\end{tabular}
\caption{Benchmark dataset statistics}
\label{table:dataset_stats}
\end{table}

\subsection{Training and Evaluation Protocol}
We compared the performance of our proposed mitigation strategy, \textit{temporal guided temperature scaled smoothing} (TGTSS), with other baseline mitigation strategies such as ensembling averaging, L2 regularization, uniform label smoothing, SWA and ensembling. We trained 50 independent models with the same hyper-parameters for each mitigation strategy using different random initialization seeds. We reported the mean ± std. dev domain classification accuracy for the Massive dataset and mean ± std. dev intent classification accuracy for the Clinc150 dataset. For both the datasets, we also reported the percentage reduction in $LE_m$ when compared to the control baseline over 50 independent model runs for all the utterances as well as for high label entropy utterances whose label entropy was over 0.56 in the control baseline. For each method, we computed the sum of $LE_m$ over all the $N$ utterances in the test set as $\sum_{i=1}^{N} LE_{m_i}$. The $\Delta LE_{m}$ is then computed as the percentage reduction among these values for each method and the control baseline. We do similar computations for $\Delta LE_{s}$ in Table 4.

The $LE_m$ value 0.56 for an utterance indicates that if the utterance was assigned to 2 different labels over 50 independent model runs, then its membership is split 75\%-25\% between the two labels. A lower value of label entropy indicates better model robustness and consequently, lower local instability. An utterance will have $LE_m = 0$ if it is consistently predicted to be the same label across 50 independent model runs. All the results for both the benchmark datasets have been reported on an unseen holdout set. A model having high overall accuracy and low label entropy is usually preferred.

\subsubsection{Hyper-parameters}
In our empirical analyses, all the models across different mitigation strategies were trained using the ADAM~\cite{Kingma_adam} optimizer with a learning rate of 0.0001. For both the benchmark datasets, all the models were trained for 5 epochs with a batch size of 256. For the control baseline with L2 regularization, we selected a weight decay value of 0.001. For the ensemble baseline, we selected $N$ as 200 i.e. the pre-temperature scaled "soft" labels were computed after uniformly averaging outputs from 200 independent models for each utterance in the training set. In the uniform label smoothing mitigation strategy, we used $\alpha$ as 0.5 for the Clinc150 dataset and  $\alpha$ as 0.1 for the Massive dataset. For SWA, we equally averaged the model weights after the last 2 epochs. For experiments using temporal guided temperature scaled smoothing on the Clinc150 dataset, we used $N$ as 200 where as for the Massive dataset, we set $N$ as 180. This indicates that model outputs after first 200 training or optimization steps were recorded for the Clinc150 dataset and uniformly averaged for each utterance  before temperature scaling. Similarly, for the Massive dataset, model outputs were recorded after 180 training steps. For both the ensemble guided and temporal guided temperature scaled smoothing mitigation strategies, we set the temperature $T$ at 0.5.

\begin{table*}[!ht]
\small
\centering
\begin{tabular*}{0.94\textwidth}{l|ccc|ccc}
\toprule
                  & \multicolumn{3}{c}{\textbf{Massive}}                 & \multicolumn{3}{c}{\textbf{Clinc150}} \\\midrule
Methods           & Accuracy(\%)                         & $\Delta LE_{m} (\%) \uparrow$  & \% of $E_b$   & Accuracy(\%)      & $ \Delta LE_{m} (\%) \uparrow$ & \% of $E_b$ \\ \midrule
Control baseline  & 90.6 ± 0.6                           & -                         & -        & 95.1 ± 0.8          & -              & -\\ 
Ensemble baseline ($E_b$) & 91.3 ± 0.5                           & 34.5                      & -  & 95.4 ± 0.6 & 31.1 & -\\\midrule
L2 Regularization & 90.3 ± 0.5                           & -2.3                      & -7            & 94.9 ± 0.7          & -0.6 & -2       \\
SWA               & 91.0 ± 0.5                           & 17.6                      & 51           & 95.2 ± 0.7          & 7.3 & 23          \\
Label Smoothing   & 90.8 ± 0.5                           & 5.7                       & 17           & 95.2 ± 0.8          & 6.1 & 20         \\\midrule
TGTSS (Ours)      & \textbf{91.3 ± 0.6}                  & \textbf{31.4} & \textbf{91}           & \textbf{95.3 ± 0.8}          & \textbf{26.7} & \textbf{86}        \\\bottomrule
\end{tabular*}
\caption{Reduction of multi-run entropy $LE_m$ across 50 independent model runs for different methods. $\Delta LE_{m} (\%)$ is calculated as percentage reduction between the sum of per-utterance $LE_m$ for each method and that of the control baseline. A higher percentage indicates greater reduction in $LE_m$ over control baseline and thus better performance. The values for \% of $E_b$ indicates the reduction in $LE_m$ as a percentage of the gold standard ensemble baseline. A negative sign in label entropy reduction indicates an increase in $LE_m$. Our method TGTSS shows the best results among the competing methods, coming within 91\% of gold standard ensemble baseline. 
}
\label{table:multirun_LE}
\end{table*}

\subsection{Results}
We compared the proposed mitigation strategy with other baselines described in Section 4.1. We highlight the effectiveness of our proposed local instability metric, \textit{label entropy}, in capturing local instability over 50 independent model runs as well as a single model run. 

\subsubsection*{Ensemble is the best mitigation strategy}  
In our empirical analyses, we found that ensemble baseline is often the best performing mitigation strategy in terms of both model accuracy and $LE_m$ for both the benchmark datasets(Table~\ref{table:multirun_LE}). 

\subsubsection*{TGTSS is comparable to ensembing at a fraction of computation cost}

We found that TGTSS is able to recover about 91\% of the performance of ensembling in the multi-run experiments. TGTSS trains only one teacher-student pair and drastically reduces the computational cost of ensembling. Hence, it is much more feasible to deploy TGTSS in production NLU systems. We also found that TGTSS is significantly better than model-centric local instability mitigation strategies such as SWA and L2 regularization. 

However, as mentioned in Section \ref{sec:tgtss}, TGTSS computes ``soft'' labels across multiple optimization steps which leads to multiple inference cycles. In our experiments, we ran inference after each optimization step once the model is no longer in the transient training state. However, it may be possible to further reduce the number of inference cycles by running inference after every $X$ optimization steps and this is left for future studies. 

\begin{table}[]
\centering
\begin{tabular}{p{3cm}|cc}
\toprule
                & \multicolumn{2}{c}{$\Delta LE_s$ (\%) $\uparrow$ } \\\midrule
Methods         & Massive                & Clinc150 \\\midrule
Label Smoothing & 37.9                   & 40.5 \\
Ensemble baseline           & \textbf{55.5}          & \textbf{61.7} \\\midrule
TGTSS (Ours)    & 53.4                   & 55.9 \\\bottomrule             
\end{tabular}
\caption{Empirical analyses highlights Temporal guided temperature scaled smoothing (TGTSS) reduces $LE_s$ with respect to the single run control baseline model across different optimization steps when a single model is trained. $\Delta LE_s$ (\%)  is computed as percentage reduction between the sum of per-utterance $LE_s$ for each method and that of the control baseline. A $-ve$ sign indicates an increase in label entropy over the control baseline.}
\label{table:singlerun_LE}
\end{table}

\begin{figure*}[!htbp]
\centerline{\includegraphics[width=16cm,height=7cm]{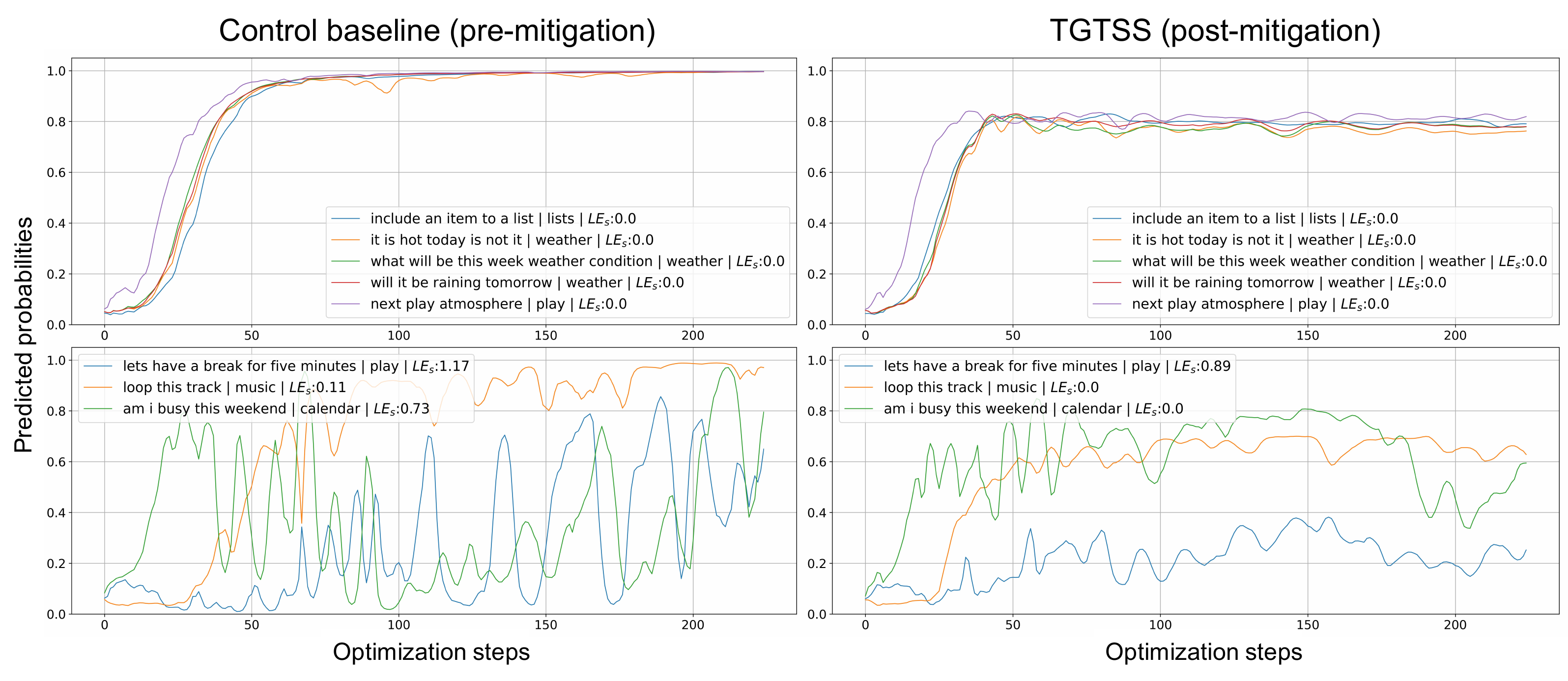}}
\caption{Training trajectories between pre-mitigation and post-mitigation stages show that TGTSS was able to significantly reduce the variability of raw confidence scores on the gold labels as well as reduce model churn in Massive dataset. [Top] shows some utterances where the model predictions are stable (no label switching), [Bottom] shows some utterance where TGTSS significantly reduced model churn as measured using $LE_s$.}
\label{figure:pre_post_mitigation}
\end{figure*}

\subsubsection*{Efficacy of single run label entropy ($LE_s$) as a local instability metric}
In Table~\ref{table:multirun_LE}, we demonstrated how TGTSS is able to reduce local instability in terms of our proposed metric $LE_m$ over multiple independent runs of the model and recover 91\% of the performance of ensembling. We propose $LE_s$ as a more practical metric for local instability. We show that TGTSS is still able to recover more than 90\% of the performance of ensembling for the Clinc150 and the Massive datasets (Table~\ref{table:singlerun_LE}). For high $LE_m$ utterances in the control baseline, TGTSS was able to considerably reduce $LE_s$ (Appendix Table~\ref{appendix_table:LE_s_high_entropy_samples}).

In figure \ref{figure:pre_post_mitigation} we observe that TGTSS significantly reduces variation in prediction scores compared to the control baseline. In the top panels we show utterances that are easy to learn and the classifier converges to the gold label within 2 epochs. In bottom panels, we show examples that exhibit high variation in prediction scores through the training process, and consequently, high $LE_s$. After mitigation by TGTSS, the bottom right panel shows the significant reduction in prediction score variation and $LE_s$. Figure \ref{fig:Massive_pre_post_mitigation_le} in Appendix shows more examples of reduction in $LE_s$ over the course of training. 


\subsubsection*{Global label smoothing is not as effective}
In our empirical analyses, we found that uniform label smoothing reduces local instability by 7-9\% compared to the control baseline but falls short of ensembling and TGTSS. Label smoothing involves computing a weighted mixture of hard targets with the uniform distribution where as both ensembling and TGTSS uses the model's average predictions over multiple runs and multiple optimization steps, respectively. Tuning the smoothing factor ($\alpha$) did not improve model stability in terms of label entropy. 

\subsubsection*{Importance of temperature scaling for TGTSS}
We conducted ablation studies to understand how temperature scaling affects the performance of TGTSS. Temperature scaling uses a parameter $T<1$ for all the classes to scale the uniformly averaged predictions. We found that the proposed methodology reduces label entropy by 17.5\% over the control baseline without temperature scaling for the Massive dataset on the validation set (31.5\% reduction with temperature scaling). This also indicates that temporal uniform averaging is independently able to significantly reduce label entropy.


\vspace{-0.25cm}
\section{Conclusion}
\vspace{-0.15cm}

In this work, we study the problem of model instability/churn in deep neural networks in the context of large scale NLU systems. Assigning different labels to the same training data over multiple training runs can be detrimental to many applications based on DNNs. We notice that the instability of model predictions are non-uniform over the data, hence we call it local instability. We propose a new metric, \textit{label switching entropy}, that is able to quantify model instability over multi-runs as well as a single training run. We also introduce \textit{Temporal Guided Temperature Scaled Smoothing} that reduces model churn by a considerable margin. We show in experiments that TGTSS is able to recover up to 91\% of the performance of ensembling at a fraction of computational cost for training and storing, thereby providing a viable alternative to ensembling in large scale production systems. Future directions of research include expanding our analysis to multi-modal data and further dissecting the root causes behind local model instability.

\vspace{-0.25cm}
\section*{Limitations}
\vspace{-0.15cm}

Even though our proposed methodology, TGTSS, was able to significantly reduce model instability, there is still a gap in performance with the gold standard ensembling techniques. More work needs to be done to bridge this gap. In our empirical analysis, we used two open source datasets, Massive and Clinc150. Both these datasets are small and may not represent the complexity in real world production datasets which may contain substantially large noise. In our proposed methodology, we train a pair of models successively, a teacher and a student, which is significantly better than ensembling in terms of computational cost. However, this setup may still be challenging in many sophisticated real world production NLU systems. More work needs to be done to reduce the computational complexity of training and inference for these systems. 

\vspace{-0.25cm}
\section*{Ethics Statement}
\vspace{-0.15cm}
The authors foresee no ethical concerns with the
research presented in this work.

\vspace{-0.25cm}
\section*{Acknowledgement}
\vspace{-0.15cm}
The authors would like to thank the anonymous reviewers and area chairs for their
suggestions and comments.

\small
\bibliographystyle{acl_natbib}
\bibliography{model_variance.bib}

\appendix

\section{Appendix}
\label{sec:appendix}

\subsection{Variance confidence plots}

\begin{figure}[htb!]
 \centering
  \includegraphics[width=\columnwidth]{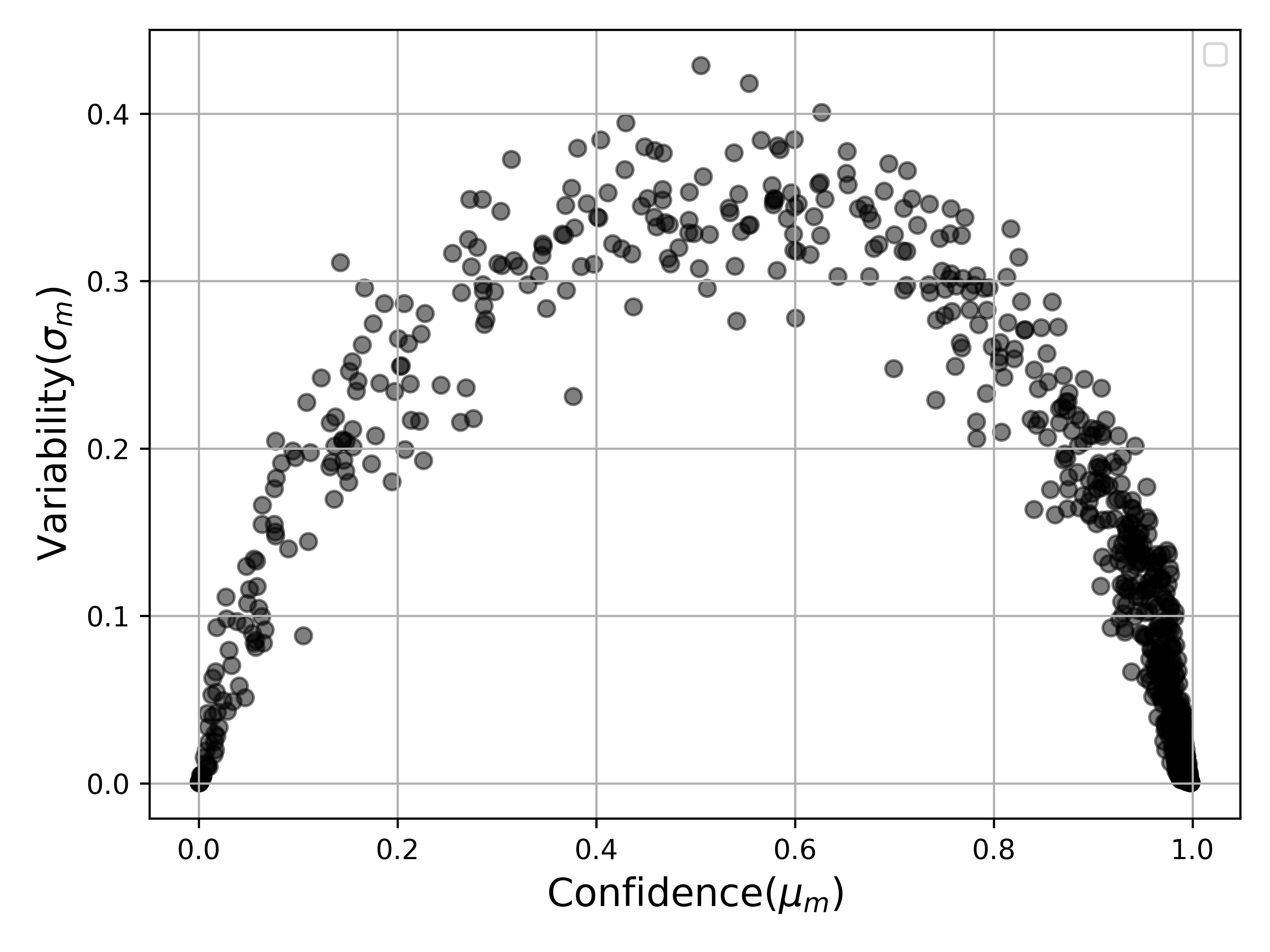}
    \caption{Plot of multi-run confidence($\mu_m$) and standard deviations($\sigma_m$) of prediction scores for Massive data (validation dataset), from the domain classifier model}
    \label{fig:Massive_mu_vs_sigma}
\vspace{-0.25cm}
\end{figure}

\begin{figure}
    \centering
    \includegraphics[width=\columnwidth]{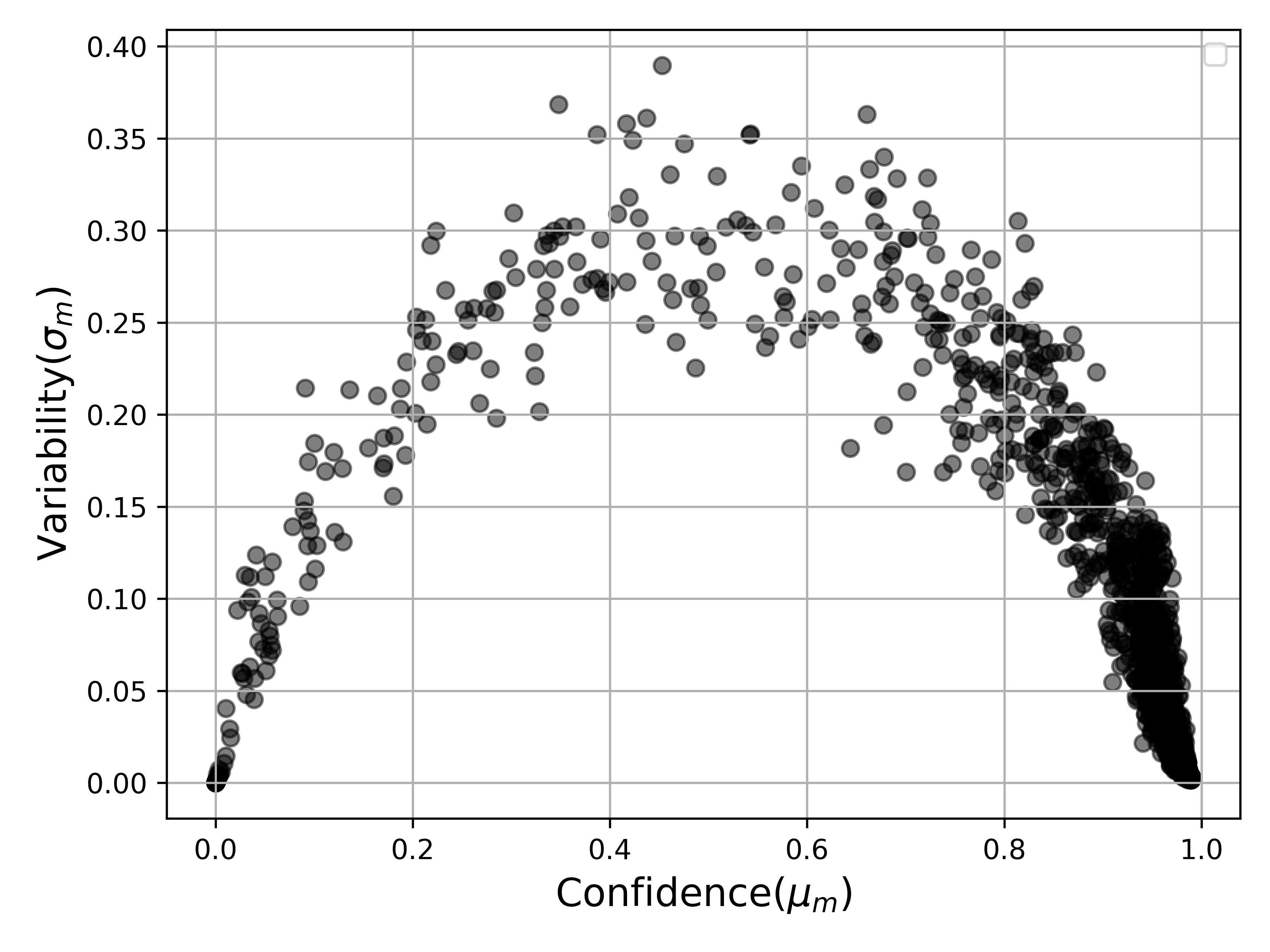}
    \caption{Plot of multi-run confidence($\mu_m$) and standard deviations($\sigma_m$) of prediction scores for Clinc150 data (validation dataset), from the intent classifier model}
    \label{fig:Clinc150mu_vs_sigma}
\vspace{-0.25cm}
\end{figure}

We have plotted the mean confidence and the variance of utterances in the validation dataset for both the Massive (Figure ~\ref{fig:Massive_mu_vs_sigma}) and Clinc150 (Figure ~\ref{fig:Clinc150mu_vs_sigma}) datasets. From our analysis, we see that there are utterances that exhibit high variance and medium confidence (around 0.5) which often leads to predicted label flips or model churn over multiple training runs of the model. We also see that there are utterances that possess low confidence corresponding to the gold label and has very low variance. These utterances are probably annotation errors. The bulk of the utterances have high confidence on average corresponding to the gold label and low confidence which signifies that the model predictions are mostly consistent on these utterances.

\subsection{Relationship between $LE_s$, $LE_m$ and $\mu_m$}

\begin{figure}[htb!]
 \centering
  \includegraphics[width=\columnwidth]{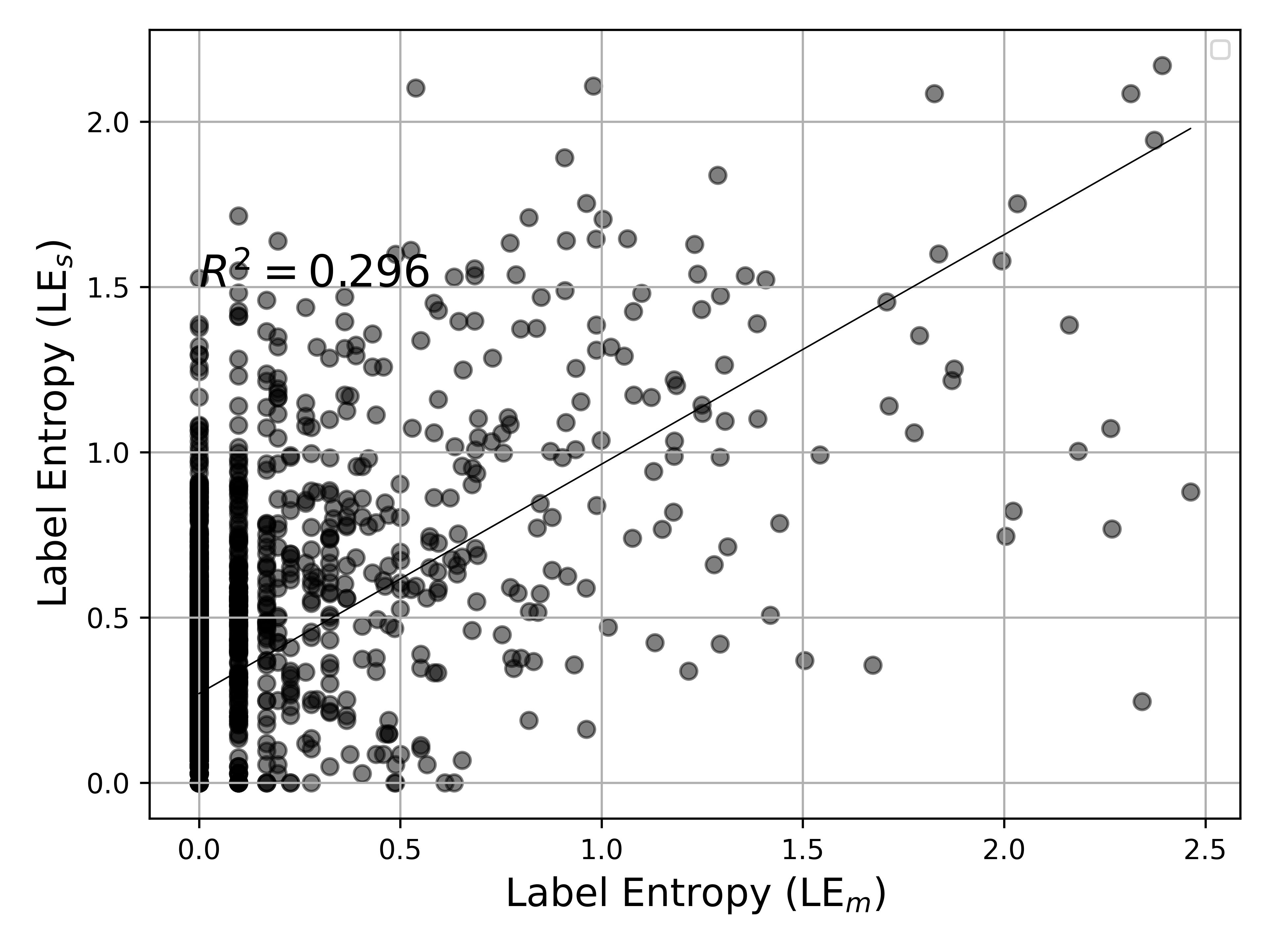}
    \caption{$LE_s$ vs $LE_m$ for Clinc150 dataset (validation set) shows a strong linear relationship. Each data point is an utterance with $LE_s^{(i)}$ vs $LE_m^{(i)}$ values. Zero entropy corresponds to utterances with confidence scores close to 1 for a class with very low variability.}
    \label{fig:Clinc150les_vs_lem}
\vspace{-0.25cm}
\end{figure}

\begin{figure}[htb!]
    \centering
    \includegraphics[width=\columnwidth]{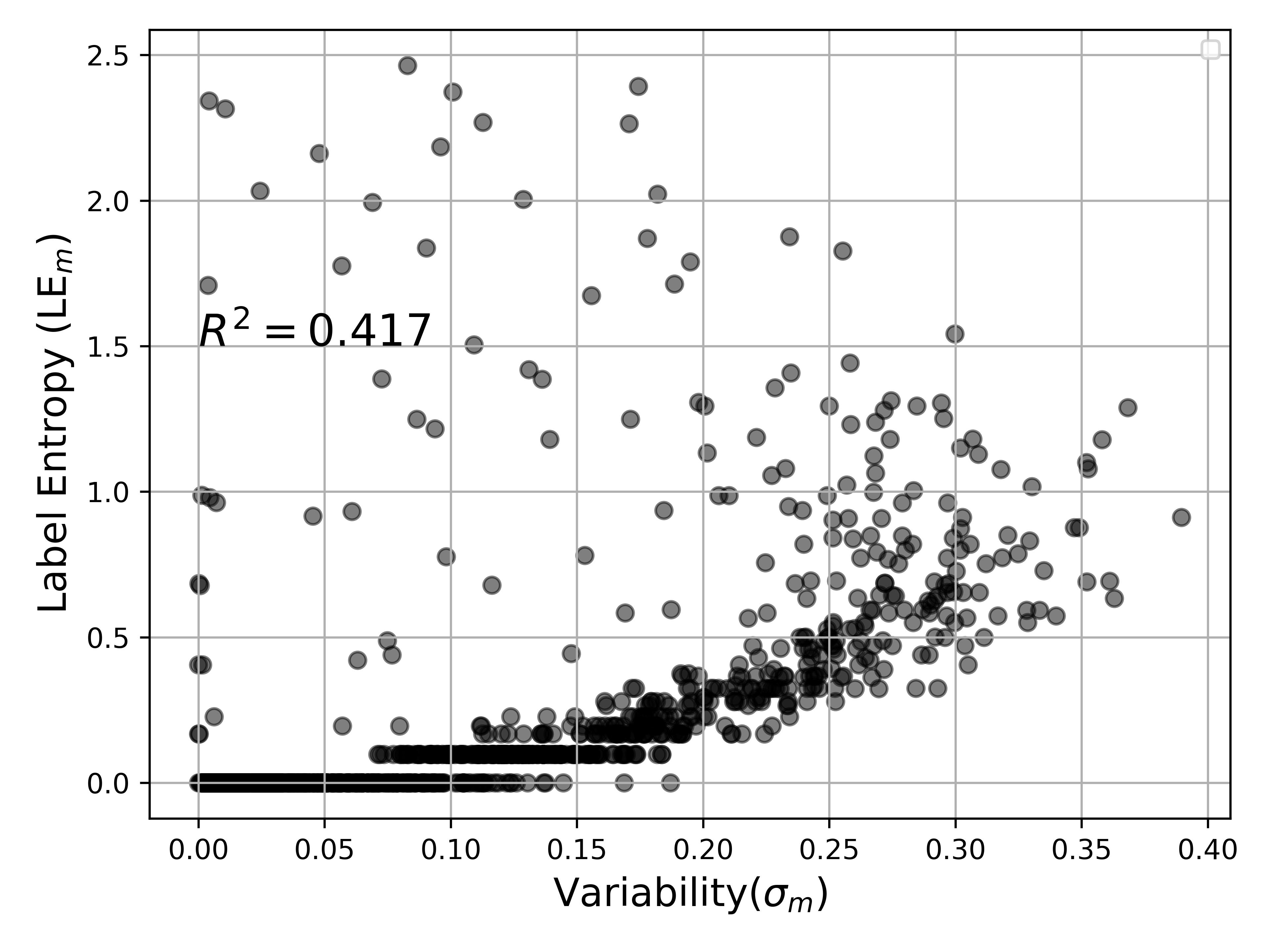}
    \caption{$LE_m$ vs $\sigma_m$ for Clinc150 dataset (validation set) shows a strong linear relationship. Each data point is an utterance with $LE_m^{(i)}$ vs $\sigma_m^{(i)}$ values.}
       \label{fig:Clinc150le_vs_sigma}
\vspace{-0.25cm}
\end{figure}

As shown earlier in the massive dataset, there is a strong relationship between $LE_m$ and $\mu_m$. We observe a similar trend in the Clinc150 dataset as well (Figure ~\ref{fig:Clinc150le_vs_sigma}). We also observe a similar relationship between single run and multiple run label entropy ($LE$) for Clinc150 dataset (Figure ~\ref{fig:Clinc150les_vs_lem}). This finding supports our analysis that label entropy is a suitable proxy for model churn.

\subsection{$LE_m$ \& $LE_s$ reduction for high entropy samples}
We computed the percentage reduction in $LE_m$ and $LE_s$ post mitigation for utterances that have high $LE_m$ in the control baseline. In our empirical studies, we showed that TGTSS was able to considerably 
reduce $LE_m$ and $LE_s$ across multi-run and single-run experiments when compared to the gold standard ensembling (Appendix Tables ~\ref{appendix_table:LE_m_high_entropy_samples},~\ref{appendix_table:LE_s_high_entropy_samples}).

\begin{table}[!ht]
\centering
\begin{tabular}{p{2.8cm}|c|c}
\toprule
                & \multicolumn{2}{c}{$\Delta LE_m$ (\%) $\uparrow$ } \\\midrule
Methods         & Massive                & Clinc150 \\\midrule
Control baseline   & - & - \\
Ensemble baseline              & 27.4 & 24.1 \\\midrule
L2 Reg.            & 3.8 & 4.2 \\
SWA                & 11.3 & 4.3 \\
Label Smoothing    & 5.4 & 8.1 \\\midrule
TGTSS (Ours)       & \textbf{26} & \textbf{22.4} \\\bottomrule             
\end{tabular}
\caption{Empirical analyses highlights TGTSS reduces $LE_m$ for high $LE_m$ samples of the control baseline by a considerable margin in multi-run experiments. The column $\Delta LE_{m} (\%) \uparrow$ is computed as percentage reduction between the sum of per-utterance $LE_m$ for each method and that of the control baseline. A higher value indicates greater reduction in $LE_m$ over control baseline.}
\label{appendix_table:LE_m_high_entropy_samples}
\end{table}

\begin{table}[!htbp]
\centering
\begin{tabular}{p{2.8cm}|c|c}
\toprule
& \multicolumn{2}{c}{$\Delta LE_s$ (\%) $\uparrow$ } \\\midrule
Methods         & Massive                & Clinc150 \\\midrule
Label Smoothing & 14.9 & 20.7 \\
Ensemble baseline & 36.4 & 40.7 \\\midrule
TGTSS (Ours) & \textbf{31.5} & \textbf{33.6} \\\bottomrule             
\end{tabular}
\caption{Empirical analyses highlights TGTSS reduces $LE_s$ for high $LE_m$ samples of the control baseline by a considerable margin in single-run experiments. The column $\Delta LE_s$ (\%) $\uparrow$ is computed as percentage reduction between the sum of per-utterance $LE_s$ for each method and that of the control baseline.}
\label{appendix_table:LE_s_high_entropy_samples}
\end{table}

\subsection{Label entropy over optimization steps}
\begin{figure*}[!htbp]
\centerline{\includegraphics[width=16cm,height=7cm]{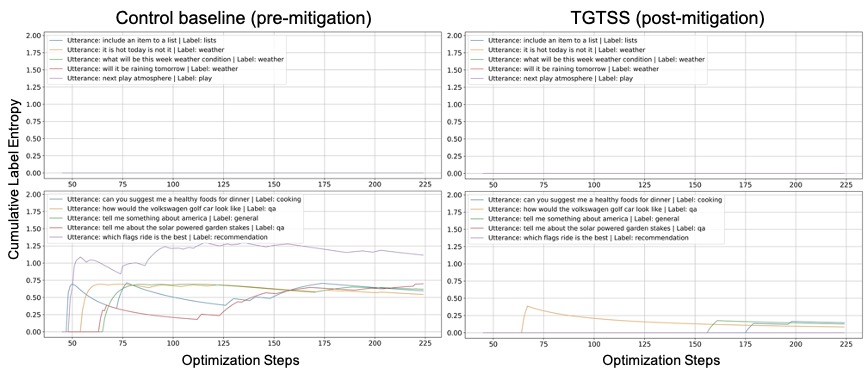}}
\caption{Training trajectories between pre-mitigation and post-mitigation stages show that TGTSS was able to significantly reduce label entropy as the model is trained. [Top] shows some utterances where the model predictions are stable as label entropy is always 0, [Bottom] shows some utterance where TGTSS significantly reduced model churn as measured using $LE_s$.}
\label{fig:Massive_pre_post_mitigation_le}
\end{figure*}

We have used $LE_s$ as a suitable proxy for $LE_m$. In Figure ~\ref{fig:Massive_pre_post_mitigation_le}, we provide empirical evidence that our proposed methodology, TGTSS, was able to reduce label entropy as the model is trained over multiple optimization steps. We computed cumulative label entropy till optimization step $T$ and observed that as the model was being trained, the label entropy of some of the utterances dropped closer to 0.  

\end{document}